\documentclass[letterpaper, 10 pt, journal, twoside]{IEEEtran}

\IEEEoverridecommandlockouts

\usepackage{graphicx}
\usepackage{amsmath}
\usepackage{amssymb}
\usepackage{cite}
\usepackage{tabularx}
\usepackage{booktabs}
\usepackage{placeins}
\usepackage[dvipsnames]{xcolor}
\usepackage[colorlinks=false,urlcolor=blue]{hyperref}

\begin{document}
\title{Morphology-Aware Graph Reinforcement Learning for Tensegrity Robot Locomotion}


\markboth{Accepted for publication in IEEE Robotics and Automation Letters, 2026}
{Zhang \MakeLowercase{\textit{et al.}}: Morphology-Aware Graph RL for Tensegrity Robot Locomotion}

\author{
Chi Zhang, Mingrui Li, Wenzhe Tong, and Xiaonan Huang
        \thanks{The authors are with the Robotics Department, University of Michigan, Ann Arbor, MI, 48109, USA.
        {\tt\footnotesize\{zhc, cldflpr, wenzhet, xiaonanh\}@umich.edu}}
}

\maketitle

\begin{abstract}
Tensegrity robots combine rigid rods and elastic cables, offering high resilience and deployability but at the same time posing major challenges for locomotion control due to their underactuated and highly coupled dynamics. This paper introduces a morphology-aware reinforcement learning framework that integrates a graph neural network (GNN) into the Soft Actor-Critic (SAC) algorithm. By representing the robot's physical topology as a graph, the proposed GNN-based policy captures coupling among components, enabling faster and more stable learning than conventional multilayer perceptron (MLP) policies. The method is validated on a physical 3-bar tensegrity robot across three locomotion primitives, including straight-line tracking and bidirectional turning. It shows superior sample efficiency, robustness to noise and stiffness variations, and improved trajectory accuracy. Additionally, the learned policies transfer directly from simulation to hardware without fine-tuning, achieving stable real-world locomotion. These results demonstrate the advantages of incorporating structural priors into reinforcement learning for tensegrity robot control.
\url{https://tensegrity-graph-rl.github.io/}.

\end{abstract}

\section{Introduction}

\IEEEPARstart{T}{ensegrity} robots, composed of rigid rods and elastic cables in prestressed configuration, have attracted increasing interest for their unique structural advantages \cite{skelton2009tensegrity, tong2024tensegrity, mi2024design}. Their intrinsic deformability provides exceptional impact resistance, enabling survival after high drops, and allows the robot to shrink into compact forms for efficient transport and deployment \cite{shah2022tensegrity}. These features make tensegrity robots promising for operation in challenging environments, ranging from disaster response to planetary exploration. However, the very properties that give tensegrity robots their advantages also complicate locomotion control.

Locomotion in tensegrity systems is challenging because of their floating-base configuration, underactuated and cable-driven design, and distributed tension network, which together yield highly non-linear and globally coupled dynamics. Local actuation propagates throughout the structure, producing complex global deformations. Such characteristics make accurate modeling difficult and limit the effectiveness of traditional control strategies.

Recent advances in deep reinforcement learning offer a potential solution, as model-free algorithms can learn policies directly from interaction without requiring explicit dynamics models \cite{Mnih2015Human}. In particular, the Soft Actor-Critic (SAC) algorithm \cite{haarnoja2018soft} has demonstrated strong performance in continuous, high-dimensional robotic control. However, most existing approaches adopt generic multilayer perceptrons (MLPs) as policy networks, treating the robot state as an unstructured vector of observations. This design neglects the robot's morphology and intrinsic coupling, often leading to slow convergence and suboptimal final performance in tensegrity locomotion. To address this limitation, we seek to incorporate the robot's structural priors directly into the policy representation.

\begin{figure}[t!]
    \centering
    \includegraphics[width=\linewidth]{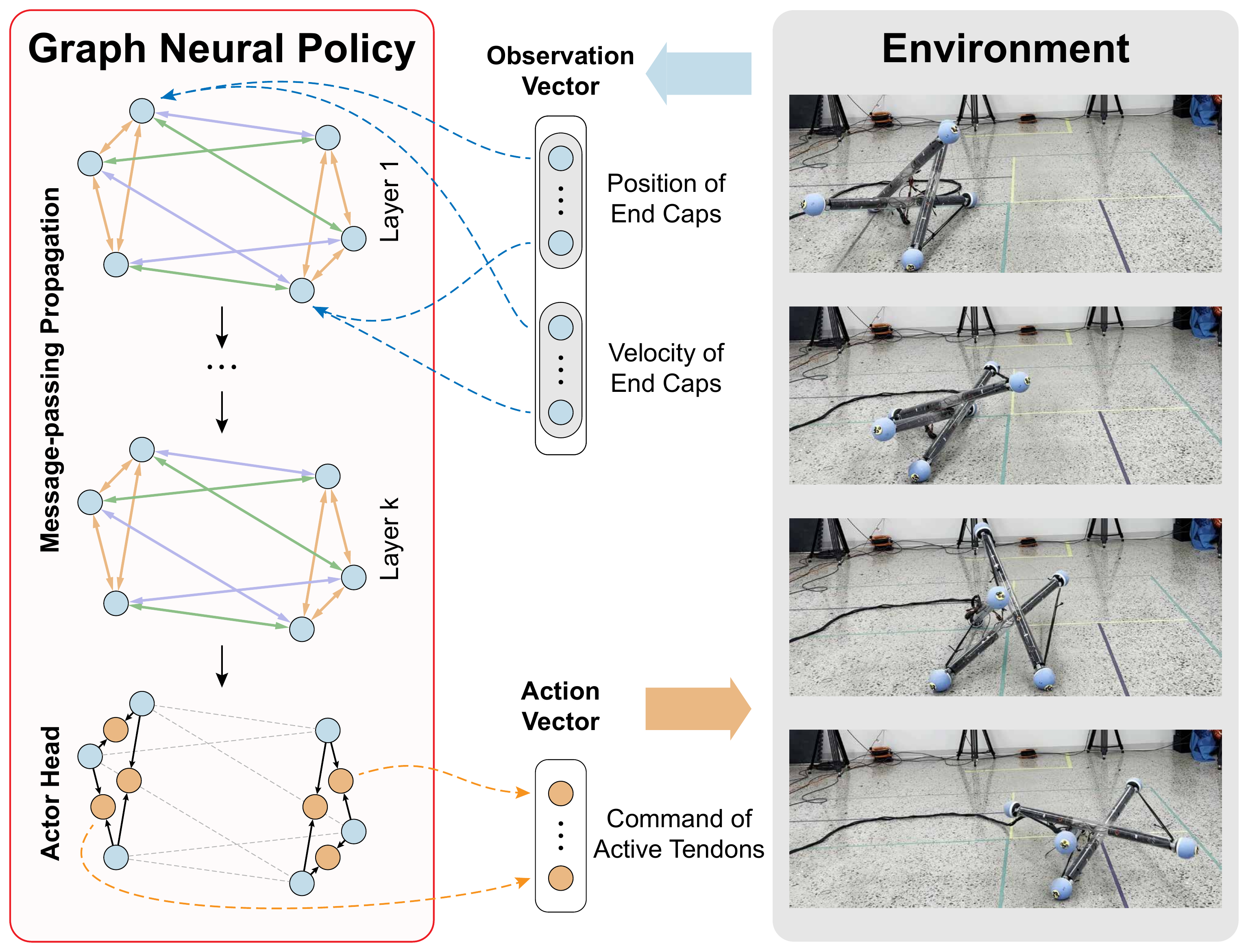}
    \caption{Morphology-aware graph reinforcement learning for tensegrity locomotion. The robot's states (end-cap positions and velocities) are encoded as node features in a graph-based policy, which propagates information along the robot's structural connections. The network outputs tendon length commands to actuate the tensegrity robot to roll forward in physical experiments.}
    \label{fig:teaser_figure}
\end{figure}

In this work, we present a morphology-aware reinforcement learning framework tailored for tensegrity robots. Our approach integrates a graph-based policy architecture into SAC, where nodes represent rod end-caps and edges represent mechanical connections via cables and rods. By explicitly encoding the robot's physical structure, the policy can capture the coupling between components, leading to more efficient and effective learning.

We evaluated the proposed framework on fundamental locomotion primitives, including straight-line tracking and in-place turning, which serve as building blocks for trajectory following. Our results demonstrate that GNN-based policies achieve higher task performance, and require fewer samples than standard MLP-based baselines. Moreover, the learned policies enable zero-shot transfer from simulation to the 3-bar tensegrity robot, resulting in robust real-world locomotion.

The contributions of this work are summarized as follows:

\begin{itemize}
    \item We propose a morphology-aware policy architecture by designing a GNN-based actor that encodes the physical topology of tensegrity robots and captures the intrinsic coupling between components.
    \item By integrating the proposed GNN actor into the SAC framework, we demonstrate substantial gains in both training efficiency and final task performance compared with conventional MLP baselines.
    \item We validate our approach on hardware by learning locomotion primitives, straight-line tracking and clockwise/counterclockwise turning, that transfer directly from simulation to physical robot without fine-tuning, demonstrating effective sim-to-real generalization.
\end{itemize}

\section{Related Work}


Early approaches to tensegrity locomotion relied on open-loop controllers with hand-designed rolling gaits \cite{paul2006design, paul2005gait, kim2015robust, kim2020rolling}, which lack feedback and are fragile under uncertainty. Model predictive control \cite{guo2021full, wang2023real} and hybrid MPC-learning methods \cite{cera2018multi} improve robustness, but faithful modeling of globally coupled tensegrity dynamics remains difficult, and MPC can be costly at scale.

These limitations motivated a shift toward data-driven methods. Reinforcement learning (RL) has achieved remarkable success across a wide range of robotic domains, with algorithms such as DDPG~\cite{lillicrap2015continuous}, PPO \cite{schulman2017proximal}, SAC \cite{haarnoja2018soft} and TD3 \cite{fujimoto2018addressing} enabling robust policies for high-dimensional continuous control. In particular, RL has powered advances in manipulation~\cite{andrychowicz2020learning}, legged locomotion \cite{Hwangbo2019Learning}, and aerial navigation \cite{Gandhi2017Learning}, and sim-to-real studies have shown that policies trained in simulation can often be deployed to hardware with minimal fine-tuning \cite{Tobin2017Domain}. These achievements highlight RL as a promising paradigm for tensegrity locomotion, where conventional model-based control struggles with complexity and modeling inaccuracies.

Several studies demonstrate that RL can produce transferable locomotion policies for tensegrity robots, even under partial observability or on rough terrain \cite{zhang2017deep, luo2018tensegrity, surovik2018adaptive, surovik2021adaptive}. While these results confirm that RL can produce robust and versatile behaviors for tensegrity robots, existing approaches almost universally adopt generic MLPs to represent policies, flattening the robot's complex morphology into an unstructured input vector. This design neglects the physical coupling intrinsic to tensegrity systems and thus often results in slow training convergence and suboptimal final performance.

Recent studies suggest that embedding morphology or structural priors into learning architectures can substantially improve efficiency and generalization \cite{wang2018nervenet, xiong2023universal, suzuki2026embeddingmorphologytransformerscrossrobot, hong2022structureaware}.

Prior work, such as NerveNet \cite{wang2018nervenet}, has explored GNN-based policies for articulated rigid-body systems. In contrast, we focus on tensegrity robots, whose compliant and tension-driven dynamics require a different morphology-aware graph formulation.

This idea has also been explored across a broader range of robotic systems. Recent work has explored structure-aware reinforcement learning across diverse robotic domains, including modular and soft robots~\cite{whitman2023learningmodularrobotcontrol}.

In the context of tensegrity robots, morphology has been incorporated into graph-based models, for instance, through graph neural networks (GNNs) used for differentiable dynamics learning \cite{chen2024learning,wang2021sim2sim}. Yet, these efforts focus primarily on system identification and forward simulation, rather than policy learning for locomotion. To bridge this gap, our work introduces a morphology-aware reinforcement learning framework that integrates GNNs into SAC, enabling policies that explicitly encode the robot's physical topology.

\section{Preliminaries}
    \subsection{3-Bar Prism Tensegrity Robot Model}

    \begin{figure}[htbp]
        \centering
        \includegraphics[width=\linewidth]{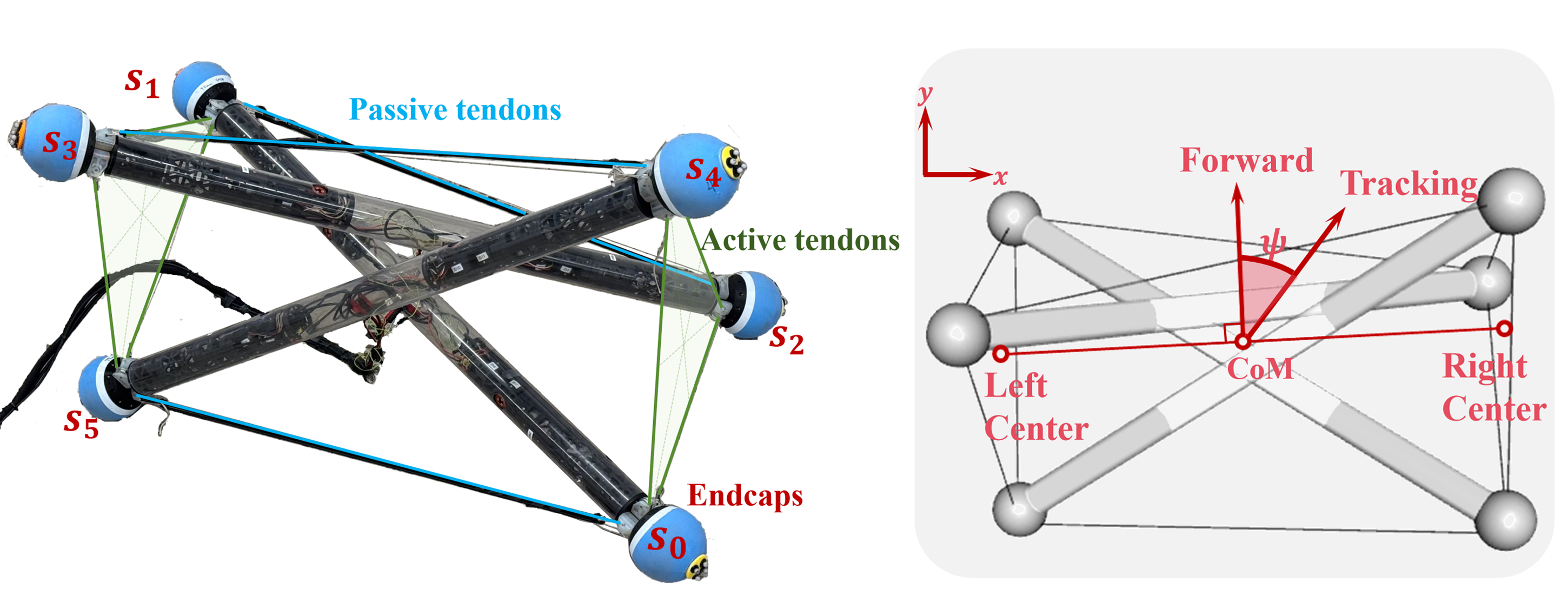}
        \caption{Physical 3-bar tensegrity robot platform and reference coordinate definitions. $\psi$ indicates the waypoint angle between forward direction and tracking direction.}
        \label{fig:hardware}
    \end{figure}
    
    The 3-bar prism tensegrity robot platform used in this study follows the design reported in~\cite{mi2024design}. This design consists of three rigid rods connected by a network of elastic tendons, forming a twisted triangular prism with two triangular faces defined as the left and right sides (Fig.~\ref{fig:hardware}). The line connecting the centroids of these faces defines the robot's lateral axis, while the forward locomotion direction lies perpendicular to this axis (Fig.~\ref{fig:hardware}, right), providing a consistent reference for orientation and motion evaluation.
    
    Each rod terminates in two end-caps, which act as structural connection points. The tendons are categorized as:

    \begin{itemize}
        \item Cross tendons (passive): connect end-caps in the same side, maintaining the global stability of the structure.
        \item Side tendons (active): connect end-caps left and right sides, and can be actuated to change length, enabling deformation and rolling motion.
    \end{itemize}

    The coordinated actuation of side tendons drives locomotion, while cross tendons preserve the tensegrity's geometry. Owing to this modular design, the robot's topology can naturally be represented as a graph: end-caps correspond to nodes, and tendons or rods define edges. This abstraction forms the foundation of the graph-based policy architecture.
    


    \begin{figure*}[t]
        \centering
        \includegraphics[width=\textwidth]{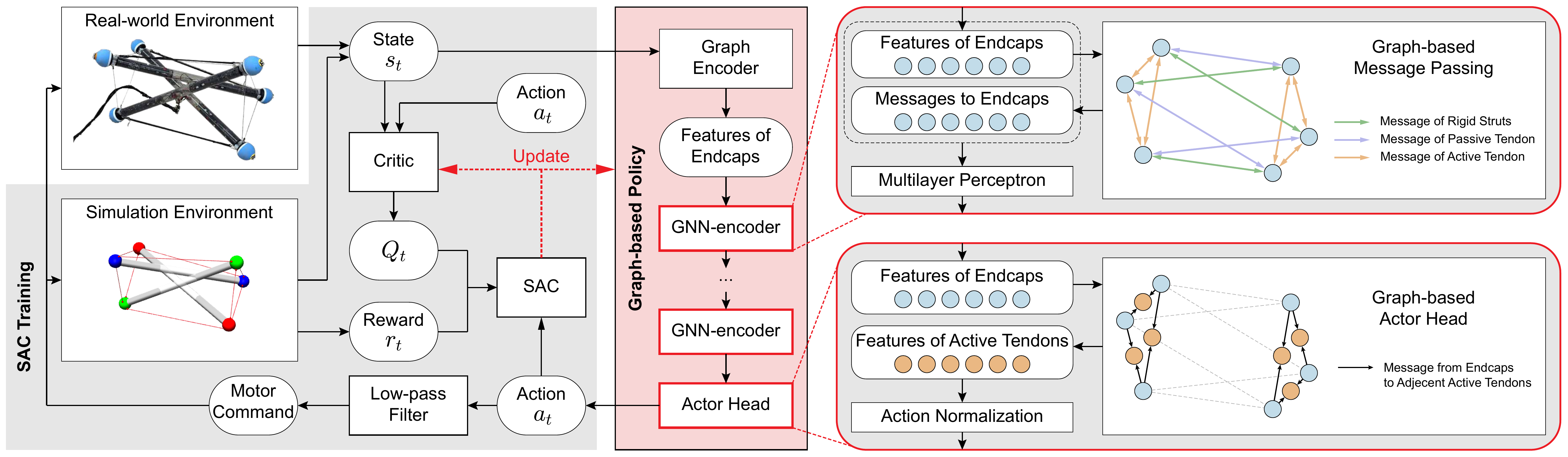}
        \caption{Overview of the proposed morphology-aware GNN-SAC framework for tensegrity robot locomotion. The Soft Actor-Critic (SAC) algorithm integrates a graph neural network (GNN)-based policy that encodes the robot's topology via message passing among end-cap nodes. The actor generates tendon length commands based on structured observations, enabling morphology-aware learning in both simulation and real-world environments.}
        \label{fig:pipeline}
    \end{figure*}

\subsection{Soft Actor-Critic Background}
\label{sec:sac_background}

    We formulate tensegrity locomotion control as a Markov Decision Process $(\mathcal{S}, \mathcal{A}, \mathcal{P}, r, \gamma)$, where states collect end-cap positions and velocities and actions are active tendon length commands.

    Soft Actor-Critic (SAC) \cite{haarnoja2018soft} optimizes an entropy-regularized objective:
    \begin{equation}
        J(\pi) = \mathbb{E}_{\tau \sim \pi} \left[ \sum_{t=0}^{\infty} \gamma^t \left( r(s_t, a_t) + \alpha \mathcal{H}(\pi(\cdot | s_t)) \right) \right],
        \label{eq:sac_objective}
    \end{equation}
    where $\alpha$ trades off task return and policy entropy. The actor and twin $Q$-critics follow the standard SAC updates (soft policy gradient and soft Bellman backups, with target networks and double $Q$-learning); $\alpha$ is automatically tuned to a target entropy. We use this SAC recipe and replace the MLP actor with the morphology-aware GNN actor in Section~\ref{sec:methodology}.

\section{Methodology}
\label{sec:methodology}
    \subsection{Graph Construction}
    
    Tensegrity robots exhibit highly coupled dynamics due to their tension-based equilibrium: a local actuation propagates through the tension network and induces global structural deformation. To effectively capture this distributed behavior, we represent the 3-bar tensegrity robot as a directed graph that mirrors its physical topology.
    
    The constructed graph, denoted as $\mathcal{G} = \left( \mathcal{V}, \mathcal{E} \right)$, contains 6 vertices $\mathcal{V}$ and 24 directed edges $\mathcal{E}$. Each vertex represents a corresponding rod-endcap $V_i$, and each physical connection (rod or tendon) is modeled as a pair of directed edges $(E_{i,j}, E_{j,i})$, enabling bidirectional message passing between vertices. The resulting graph, therefore, consists of three types of edges: i) Rigid rods (6 directed edges); ii) Passive tendons (6 directed edges); iii) Active tendons (12 directed edges).
    
    Each edge encodes the physical relationship between two end caps, while vertices carry local state information. This representation allows the policy network to leverage message passing over the morphology-aware graph structure, rather than treating the robot state as an unstructured flat vector.
    
\subsection{GNN-Based Soft Actor-Critic}
    We integrate the above tensegrity graph representation into the actor network of Soft Actor-Critic (SAC). 
    This design enables the policy to explicitly capture spatial coupling among robot components through relational message passing.

    \subsubsection{Observation Encoding}
        At each timestep, the raw robot state is encoded into the vertex and edge features of the tensegrity graph. Vertex $V_i$ features include the 3D position  $p_i$ and velocity $v_i$ of the corresponding rod end-cap, augmented with task-related parameters (e.g., global motion commands for target position) that are broadcast to all vertices. Edge features encode the relative distance between the connected vertices, together with a categorical indicator of edge type (rigid rod, passive tendon, or active tendon). These structured features serve as the input to the GNN encoder, enabling the policy to reason over the robot state in a morphology-aware manner.



    \subsubsection{GNN Encoder}
        The GNN encoder consists of stacked message-passing layers operating on a fixed graph topology.
        At each layer, an edge-wise message is computed from a sender vertex $V_j$ to a receiver vertex $V_i$ as:
        \begin{equation}
            M_{i,j} = \text{MLP}_m\big(V_i, V_j, E_{i,j}\big),
        \end{equation}
        where $E_{i,j}$ denotes the feature vector associated with the edge $(i,j)$.
        Incoming messages are aggregated at each vertex via summation:
        \begin{equation}
            M_i = \sum_{j \in \mathcal{N}(i)} M_{i,j}.
        \end{equation}
        with $\mathcal{N}(i)$ denoting the neighbors of $V_i$. The vertex representation is then updated as:
        \begin{equation}
            V_i' = \text{MLP}_f\big(V_i, M_i\big),
        \end{equation}
        where $\text{MLP}_f$ maps the concatenated input to a latent embedding of fixed dimension.
        Multiple message-passing layers are stacked to propagate information across the graph, allowing the vertices to encode rich contextual information about both local states and global coupling effects.

        

    \subsubsection{Actor Head}
        The actor head predicts control commands for active tendons, which are represented as active edges in the graph.
        For each active edge $(i,j)$, the policy outputs the parameters of a Gaussian distribution:
        \begin{equation}
            (\mu_{i,j}, \log\sigma_{i,j}) = \text{MLP}_a\big(V_i, V_j, E_{i,j}\big),
        \end{equation}
        where $\mu_{i,j}$ and $\sigma_{i,j}$ denote the mean and standard deviation of the action distribution associated with edge $(i,j)$.
        Actions are sampled from the resulting Gaussian distribution and squashed by a $\tanh$ function to enforce bounded control.
        This formulation aligns the action space with the robot morphology by directly associating each control output with a corresponding active tendon.
        The sampled action $\tilde{a}_{i,j} \in (-1,1)$ is then linearly rescaled to the feasible actuation range of tendon $(i,j)$ and applied as the control command at the current timestep.

    \subsubsection{Policy Training with SAC}
        The GNN-based actor is embedded into the standard Soft Actor-Critic (SAC) framework. At each step, the encoded graph state is processed by the GNN encoders and the actor head to produce tendon length commands, which are filtered and applied as motor targets in the environment. The SAC algorithm then updates the actor and critic networks based on observed rewards and transitions, preserving the benefits of entropy-regularized reinforcement learning while leveraging morphology-aware policy representations.

    \begin{table}[t]
      \caption{SAC training hyperparameters.}
      \label{tab:sac_hyperparams}
      \centering
      \begin{tabular}{lcc}
        \toprule
        Parameter & Value & Notes \\
        \midrule
        Learning rate & $3\!\times\!10^{-4}$ & all optimizers \\
        Discount ($\gamma$) & 0.99 & reward discount factor \\
        Target update ($\tau$) & 0.005 & soft update for target critic \\
        Batch size & 128 & batch size for updates \\
        Replay buffer & $10^{6}$ & capacity \\
        Warmup steps & 256 & random actions before updates \\
        Updates / step & 1 & gradient steps per env step \\
        Entropy coef. ($\alpha$) & auto & learned via gradient descent \\
        Target entropy & $-|A|$ & proportional to action dimension \\
        Action distribution & Gaussian & squashed with $\tanh$ \\
        Log std bounds & $[-20,2]$ & clamp \\
        \bottomrule
      \end{tabular}
    \end{table}

    \begin{table}[t]
      \centering
      \caption{Actor and critic network architectures.}
      \label{tab:network_architecture}
      \begin{tabular}{lcc}
        \toprule
        Component & Setting & Notes \\
        \midrule
        \multicolumn{3}{c}{\textbf{GNN-based Actor}} \\
        \midrule
        GNN encoder & 2 & message-passing layers \\
        Hidden dim $d_h$ & 128 & all embeddings \\
        $\text{MLP}_m$ & $(2d_v+d_e)\!\rightarrow\! d_h \!\rightarrow\! d_h$ & message fn. \\
        $\text{MLP}_f$ & $(d_v+d_h)\!\rightarrow\! d_h$ & vertex update \\
        $\text{MLP}_a$ & $(2d_h+d_e)\!\rightarrow\! d_h \!\rightarrow\!2$ & $(\mu,\log\sigma)$ \\
        Activations & ReLU & all MLPs \\
        Msg aggregation & Sum & incoming messages \\
        Action output & $(\mu,\log\sigma)$ & active edges \\
        Action bound & $\tanh$ & $[-1,1]$ \\
        \midrule
        \multicolumn{3}{c}{\textbf{MLP-based Critic}} \\
        \midrule
        Q-networks & 2 & double Q-learning \\
        Input & state + action & concatenated \\
        Hidden layers & 2 & shared structure \\
        Hidden dim & 128 & per layer \\
        Activations & ReLU & \\
        Output & 1 & scalar Q-value \\
        \bottomrule
      \end{tabular}
      \vspace{0.5em}
      \footnotesize{$d_v$, $d_e$, and $d_h$ denote vertex, edge, and hidden dimensions, respectively.}
    \end{table}

\subsection{Training Formulation}

    To enable waypoint-based navigation, we design three motion primitives for the tensegrity robot: (i) straight-line tracking toward a designated target point, (ii) counterclockwise in-place turning, and (iii) clockwise in-place turning. Each primitive is implemented as a reinforcement learning task under the GNN-SAC framework, where the reward combines task progress with control regularization. All tasks are defined in the 2-D ground plane with the robot center of mass (CoM) position denoted as $\boldsymbol{p} \in \mathbb{R}^2$ and yaw angle $\psi$. For the tracking task, the observation additionally includes a tracking vector $\boldsymbol{v}_{\text{tr}} = \boldsymbol{p}_{\text{target}} - \boldsymbol{p}$, while turning tasks use only the generic robot state.


    \subsubsection{Tracking Reward}
        For the tracking task, the robot is required to move from its initial center of mass position $\boldsymbol{p}_0 \in \mathbb{R}^2$ toward a target waypoint $\boldsymbol{p}_{\text{target}} \in \mathbb{R}^2$. The target waypoint is defined with an angular deviation $\Delta \psi$ from the initial heading $\psi_0$, constrained by a maximum bound $|\Delta \psi| \leq \Delta \psi_{\text{max}}$. This constraint ensures that the target direction remains approximately aligned with the robot's initial orientation. The desired direction is given by:
        \begin{equation}
            \boldsymbol{d}_{\text{tr}} = \frac{\boldsymbol{p}_{\text{target}} - \boldsymbol{p}_0}{\| \boldsymbol{p}_{\text{target}} - \boldsymbol{p}_0 \|_2}.
        \end{equation}

        Let $\boldsymbol{p}_t$ denote the CoM position at time $t$. The CoM displacement $\boldsymbol{p}_t - \boldsymbol{p}_0$ is decomposed into the aligned component
        $d_a(\boldsymbol{p}_t) = \boldsymbol{d}_{\text{tr}}^\top (\boldsymbol{p}_t - \boldsymbol{p}_0)$
        and the lateral deviation
        $d_b(\boldsymbol{p}_t) = \| \boldsymbol{d}_{\text{tr}} \times (\boldsymbol{p}_t - \boldsymbol{p}_0) \|_2$.
        A potential function encodes both progress and alignment:
        \begin{equation}
            P_{\text{tr}}(\boldsymbol{p}_t) = \lambda_{1} d_a(\boldsymbol{p}_t) \exp\left(-\tfrac{d_b(\boldsymbol{p})^2}{2\sigma_{\text{tr}}^2}\right) - \lambda_{2} \| \boldsymbol{p}_t - \boldsymbol{p}_{\text{target}} \|_2,
        \end{equation}
        
        The tracking reward is the potential difference over a control step:
        \begin{equation}
            r_{\text{tr}} = P_{\text{tr}}(\boldsymbol{p}_{t+1}) - P_{\text{tr}}(\boldsymbol{p}_t).
        \end{equation}

    \subsubsection{Turning Reward}  
        For in-place turning, the robot is encouraged to change yaw with minimal translation. Let $\psi_t$ denote the yaw angle of the robot at step $t$ and $\boldsymbol{p}_t$ the corresponding CoM position. A potential function penalizes translational deviation from the initial position $\boldsymbol{p}_0$:
        \begin{equation}
            P_{\text{turn}}(\boldsymbol{p}) = - \lambda_{\text{turn}} \| \boldsymbol{p} - \boldsymbol{p}_0 \|_2^2
        \end{equation}
        
        The desired turning direction is represented as $d_{\text{turn}} \in \{+1,-1\}$, where $+1$ indicates counterclockwise and $-1$ indicates clockwise rotation. The turning reward is then defined as
        \begin{equation}
            r_{\text{turn}} = d_{\text{turn}} \cdot \left( \psi_{t+1} - \psi_t \right) + \left[ P_{\text{turn}}(\boldsymbol{p}_{t+1}) - P_{\text{turn}}(\boldsymbol{p}_t) \right]
        \end{equation}

    \subsubsection{Control Cost}
        For all tasks, tendon actuation is regularized by a quadratic penalty:
        \begin{equation}
            c(\boldsymbol{a}_t) = \lambda_c \sum_i (a_{t,i} - l_{0,i})^2
        \end{equation}
        
        where $a_{t,i}$ is the commanded length of tendon $i$ at step $t$, and $l_{0,i}$ its nominal balanced length.

    \subsubsection{Overall Reward} 
        The per-step reward is expressed as
        \begin{equation}
            r_t =
            \begin{cases}
                r_{\text{tr}} - c(\boldsymbol{a}_t), & \text{tracking} \\
                r_{\text{turn}} - c(\boldsymbol{a}_t), & \text{turning}
            \end{cases}
        \end{equation}
        
        These task-specific rewards provide the optimization signal for training the GNN-based SAC policy.

        In the experiments, the reward weights are set to $\lambda_1 = \lambda_2 = \lambda_{\text{turn}} = 50$ and $\lambda_c = 0.01$.

\section{Experimental Results}
    \subsection{Experimental Setup}
    Simulations use MuJoCo at 50~Hz with passive tendons modeled as linear springs ($k = 450~\mathrm{N/m}$) and low-pass filtered actions.
    Training is implemented in PyTorch on an Ubuntu~20.04 laptop (Intel i5-12500H, RTX 3070Ti). On a Jetson Orin Nano, mean inference time is 9.61~ms ($\sim$105~Hz), sufficient for onboard control.

    The hardware platform is constructed with polycarbonate rods, Dyneema active tendons, and nonlinear elastic passive polymer ropes (242--643~N/m); each active tendon uses a GIM4310 BLDC motor with 10:1 gearbox and spool, with actuation range of [0.1, 0.8]m.

    MoCap system at 100~Hz (2~mm std) streams end-cap positions to a Jetson Orin Nano (ROS); CAN at 50~Hz closes the loop to the motor controllers for real-world rollouts.

\subsection{Benchmark of Learning Performance}


    We benchmark \textbf{G-SAC} (graph policy), \textbf{M-SAC} (MLP policy), \textbf{PPO}, and \textbf{TD3} on straight-line tracking, counterclockwise turning, and clockwise turning.
    
\begin{figure}[htbp]
        \centering
        \includegraphics[width=\linewidth]{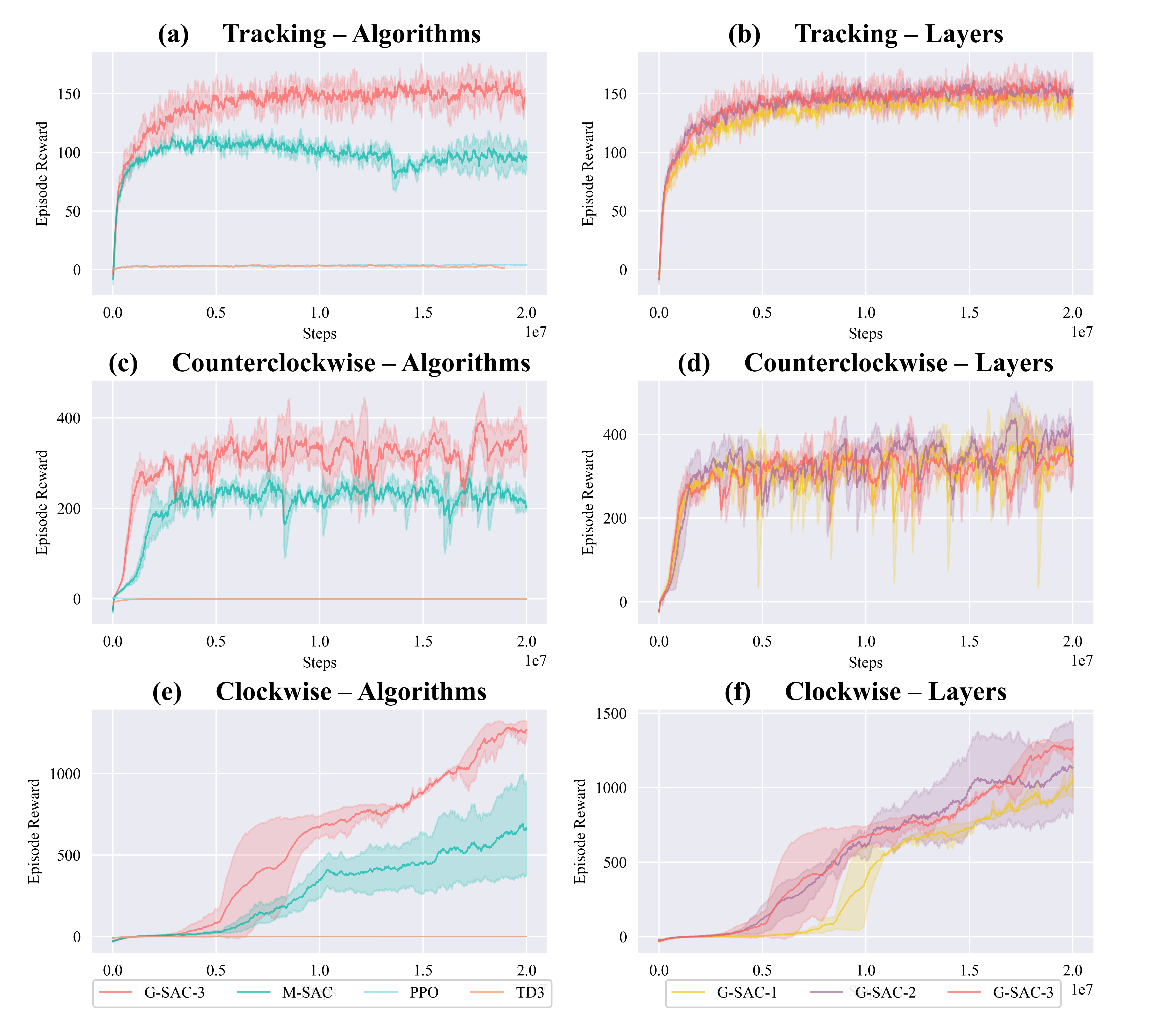}
        \caption{Benchmark of learning performance across algorithms and network depths. The proposed GNN-SAC consistently outperforms MLP-based SAC (M-SAC), PPO, and TD3 in terms of training reward and sample efficiency for all three locomotion primitives. Subplots (a,c,e) compare algorithms, while (b,d,f) analyze the effect of GNN encoder depth, showing improved performance with multi-layer message passing.}
        \label{fig:benchmarking}
    \end{figure}

    As shown in Fig.~\ref{fig:benchmarking} a, c, and e, G-SAC consistently outperforms the baselines in both reward per training step and reward per wall-clock time across all primitives. The advantage over M-SAC indicates that explicitly encoding the robot's morphology significantly accelerates policy learning, while the performance gap to PPO and TD3 reflects the efficiency of entropy-regularized off-policy actor-critic methods in high-dimensional continuous control.

    We further conducted an ablation study on the depth of the GNN encoder, comparing single-layer (\textbf{G-SAC-1}), two-layer (\textbf{G-SAC-2}), and three-layer (\textbf{G-SAC-3}) variants. As shown in Fig.~\ref{fig:benchmarking}b, d, and f, deeper encoders generally yielded better performance: G-SAC-3 achieved slightly higher final rewards than G-SAC-2, while both substantially outperformed G-SAC-1. This trend reflects the benefit of multi-hop message passing, allowing the policy to capture long-range coupling across the tensegrity structure.

    The average training speed for M-SAC, G-SAC-1, G-SAC-2, G-SAC-3 are 62.2, 39.3, 35.5 and 32.1 steps per second with same optimization, respectively. It is seen that the G-SAC still have significant advantage over M-SAC in time domain.
    

    Overall, these results establish that incorporating structural priors via GNNs improves both sample efficiency and policy quality, forming a strong baseline for the subsequent evaluations.

    \begin{figure}[htbp]
        \centering
        \includegraphics[width=\linewidth]{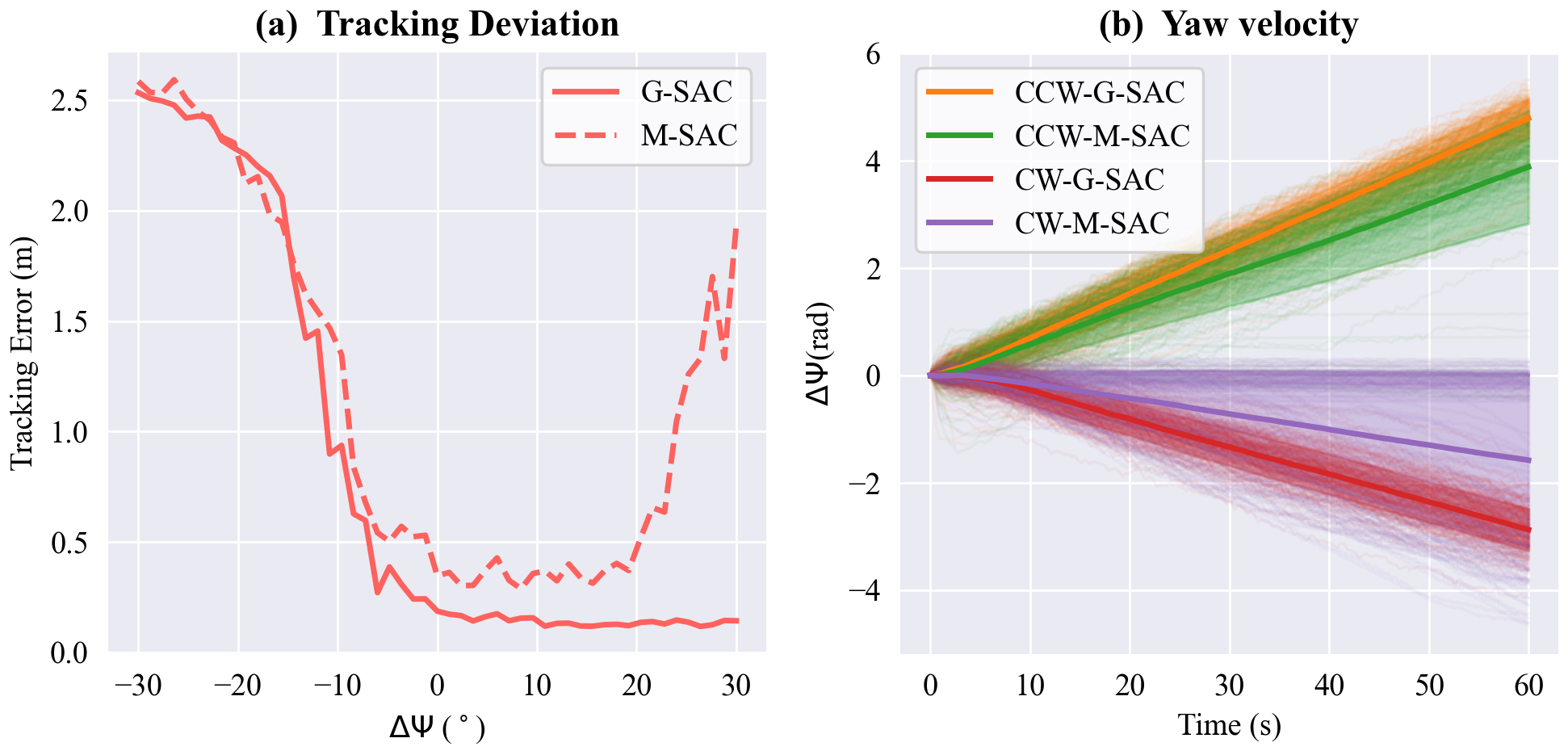}
        \caption{Simulation evaluation of learned motion primitives between Graph-based SAC (G-SAC) and MLP-based SAC (M-SAC): (a) Straight-line tracking error for different waypoint yaw angles; (b) Yaw rate and stability in bidirectional turning tasks.}
        \label{fig:tracking}
    \end{figure}
    
    \begin{figure}[htbp]
        \centering
        \includegraphics[width=1.0\linewidth]{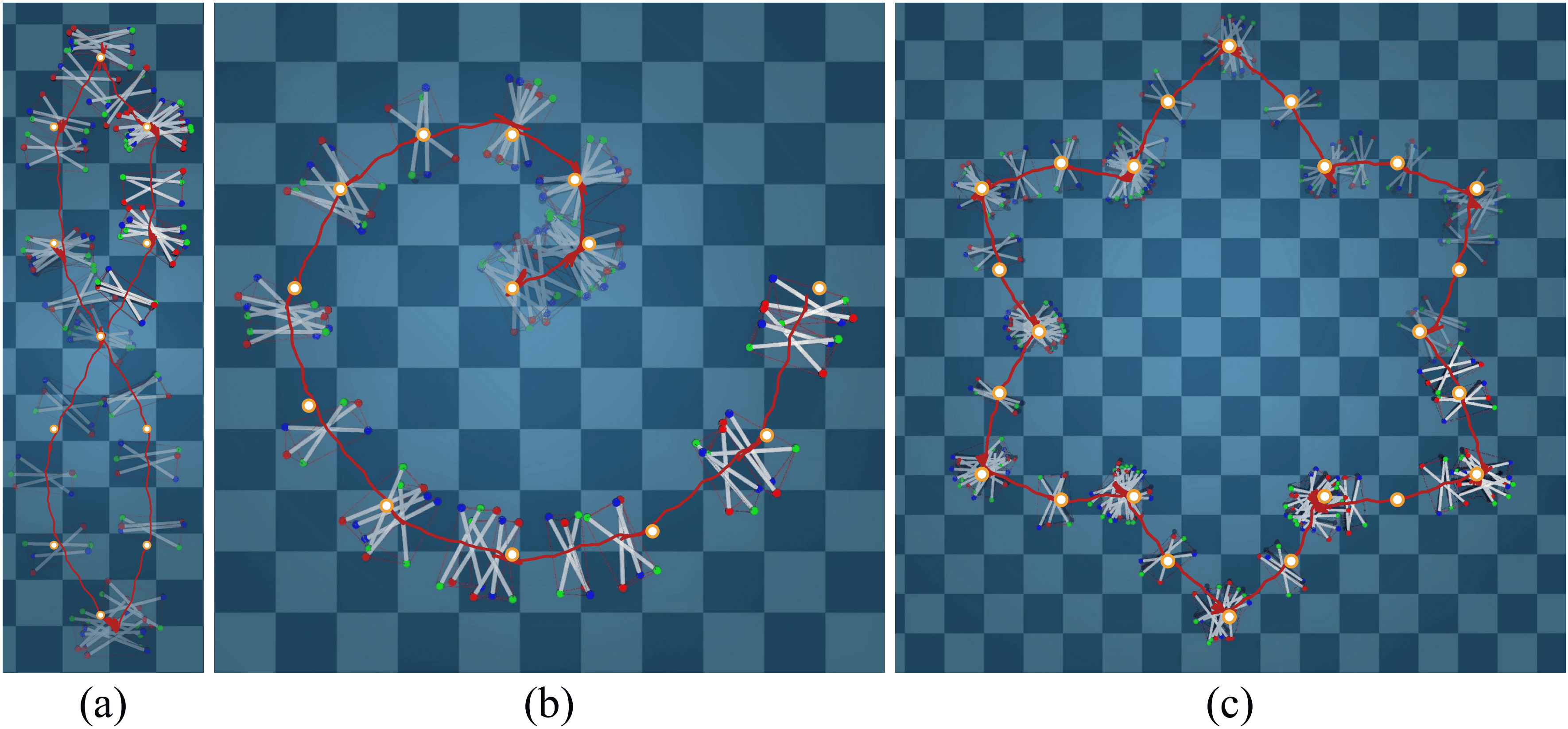}
        \caption{ Composed trajectory tracking using learned motion primitives. The robot follows an $\infty$-shaped (a), spiral (b), and six-petal flower (c) waypoint sequence by sequentially combining straight-line and turning primitives. The resulting CoM trajectory (red) closely aligns with target waypoints (orange), confirming effective motion composition and trajectory accuracy.}
        \label{fig:sim_traj}
    \end{figure}

\subsection{Motion Primitive Evaluation and Trajectory Composition}
    After training, we evaluated all three primitives in simulation under randomized initial poses.
    
    For straight-line tracking, the robot reached waypoints at several heading offsets; performance was quantified by the standard deviation of the robot's final-position distribution centered on target position.
    As shown in Fig.~\ref{fig:tracking}(a), the G-SAC policy achieved substantially lower deviation in distance than M-SAC across all waypoint directions, demonstrating improved directional accuracy and robustness.

    For in-place turning, policies were evaluated by the average yaw rate during counterclockwise (CCW) and clockwise (CW) rotations. G-SAC exhibited faster and more stable rotations compared to M-SAC as shown in Fig.~\ref{fig:tracking}(b).

    To demonstrate motion composability, the three primitives were combined for waypoint-based trajectory following. A high-level planner sequentially selected primitives according to the robot's relative pose to successive targets. As illustrated in Fig.~\ref{fig:sim_traj}, the robot successfully reached a sequence of waypoints, showing that the learned primitives can serve as reliable building blocks for higher-level navigation.

    Together, these results confirm that graph-based policies not only accelerate learning but also enable precise, stable, and composable locomotion behaviors.

    \begin{figure*}[htbp]
        \centering
        \includegraphics[width=\linewidth]{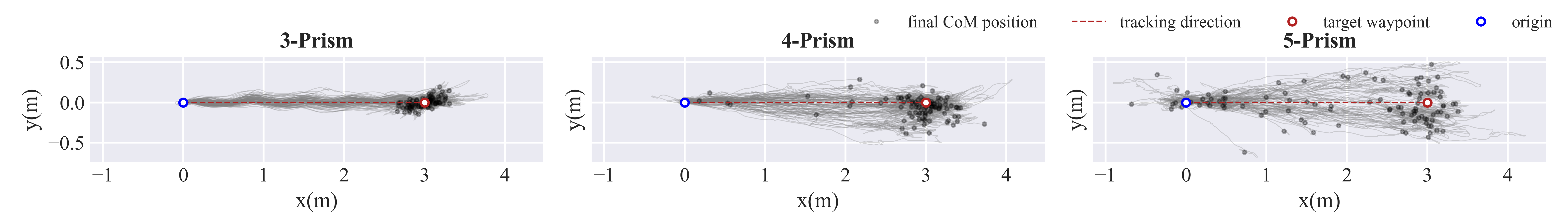}
        \caption{Cross-morphology generalization of the GNN-based policy. A policy trained on the 3-bar prism tensegrity robot is directly deployed to 4-bar and 5-bar configurations without retraining. Each subplot aggregates 100 trials. The policy achieves accurate tracking on the 3-bar system, while maintaining functional locomotion with gradually increasing deviation on 4-bar and 5-bar systems, demonstrating zero-shot generalization across morphologies.}
        \label{fig:cross_morphology_generalization}
    \end{figure*}
        
\subsection{Robustness Evaluation}
    \label{sec:robustness_eval}
    \begin{figure}[t]
        \centering
        \includegraphics[width=1.0\linewidth]{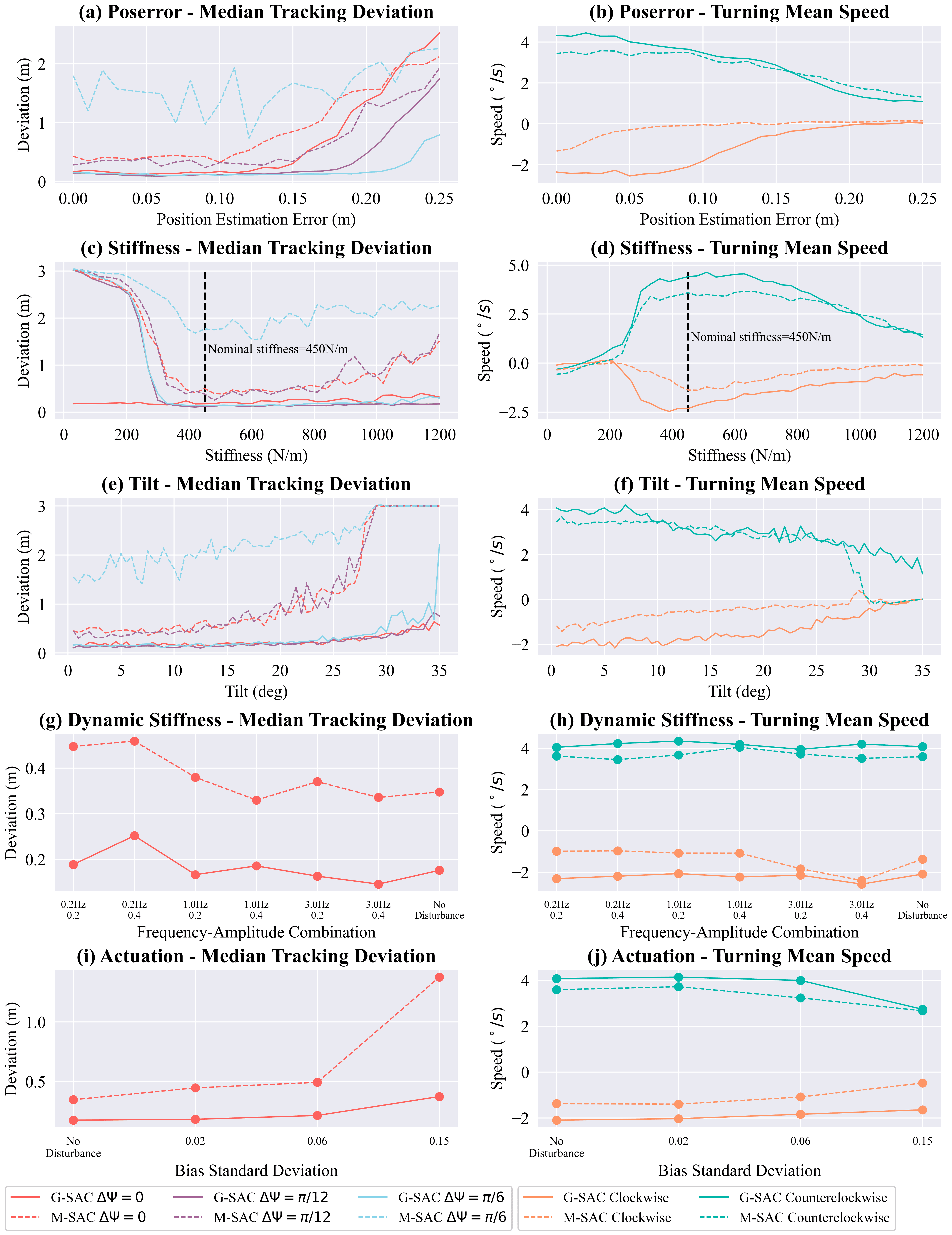}
        \caption{Robustness evaluation under model and environment perturbations. (a) (b) Cross-tendon stiffness variation, (c) (d) Observation noise, (e) (f) Ground slope, (g) (h) Dynamic stiffness variation, (i) (j) Actuation error.}
        \label{fig:robustness}
    \end{figure}
    
    To evaluate robustness under real-world-like uncertainties, we introduced controlled perturbations in simulation along the following dimensions:

    \begin{itemize}
        \item \textbf{Cross-tendon stiffness:} varied from 30-1200 N/m (nominal 450~N/m) to capture material and assembly variations.
        \item \textbf{State estimation noise:} Gaussian noise $\mathcal{N}(0, \sigma_n)$ with $\sigma_n \in [0, 0.25]$~m was added to end-cap positions to simulate degraded sensing.
        \item \textbf{Ground slope:} inclinations from $0^{\circ}$ to $35^{\circ}$ were tested using policies trained on flat terrain.
        \item \textbf{Dynamic stiffness variation:} sinusoidal stiffness perturbations with amplitudes of $\{ \pm 20\%, \pm 40\%\}$ at frequencies of $\{0.2, 1.0, 3.0\}~Hz$ around the nominal value.
        \item \textbf{Actuation error:} random bias with standard deviations $\sigma_a \in \{0.02, 0.06, 0.15\}$~m was added to control signals to simulate actuator imperfections and cable slack.
    \end{itemize}

    Fig.~\ref{fig:robustness} summarizes the trends: under observation noise up to $\sigma_n = 0.15$~m, tracking accuracy was maintained with only moderate degradation in turning speed. Turning rates preserved over half of their maximum for stiffness values between 300-800~N/m, and tracking performance remained stable within this range. On inclined planes, the robot remained stable up to about $25^\circ$ tilt. Under dynamic stiffness variations with amplitudes up to $\pm 40\%$, the robot demonstrated more robust tracking performance across all tested combinations. With actuation errors up to $\sigma_a = 0.15$~m, tracking capability was maintained while counterclockwise turning remained functional. Across all perturbations and primitives, G-SAC consistently outperformed M-SAC, exhibiting smoother trajectories and greater tolerance.

    These results indicate that embedding morphological structure in the policy network not only improves nominal performance but also enhances robustness to uncertainties critical for real-world deployment.

    

\subsection{Cross-Morphology Generalization}

    To evaluate the generalizability of the proposed policy across different robot configurations, we conduct cross-morphology experiments on prism tensegrity robots with varying numbers of bars. Specifically, a straight-line tracking policy trained on the 3-bar prism tensegrity robot is directly deployed, without retraining, to 4-bar and 5-bar prism configurations in simulation. For each morphology, 100 trials are performed with randomized initial poses.

    Fig.~\ref{fig:cross_morphology_generalization} shows that 3-bar tracking stays close to the reference, whereas 4-bar and 5-bar remain functional with increasing dispersion in final positions and trajectory variability.

    These results indicate that the proposed GNN-based policy generalizes across tensegrity morphologies in a zero-shot manner. While locomotion accuracy degrades as the structural difference from the training morphology increases, the degradation is gradual and does not lead to failure, indicating that the policy captures transferable structural priors through its graph representation.

    It is important to note that conventional MLP-based policies cannot be directly applied in this setting due to their fixed input and output dimensions, which depend on the specific robot configuration. The proposed GNN-based policy, however, naturally accommodates varying graph sizes and topologies, allowing a single trained model to generalize across different tensegrity structures without architectural modification.
    
\subsection{Sim-to-Real Transfer}
    \begin{figure}[htbp]
        \centering
        \includegraphics[width=\linewidth]{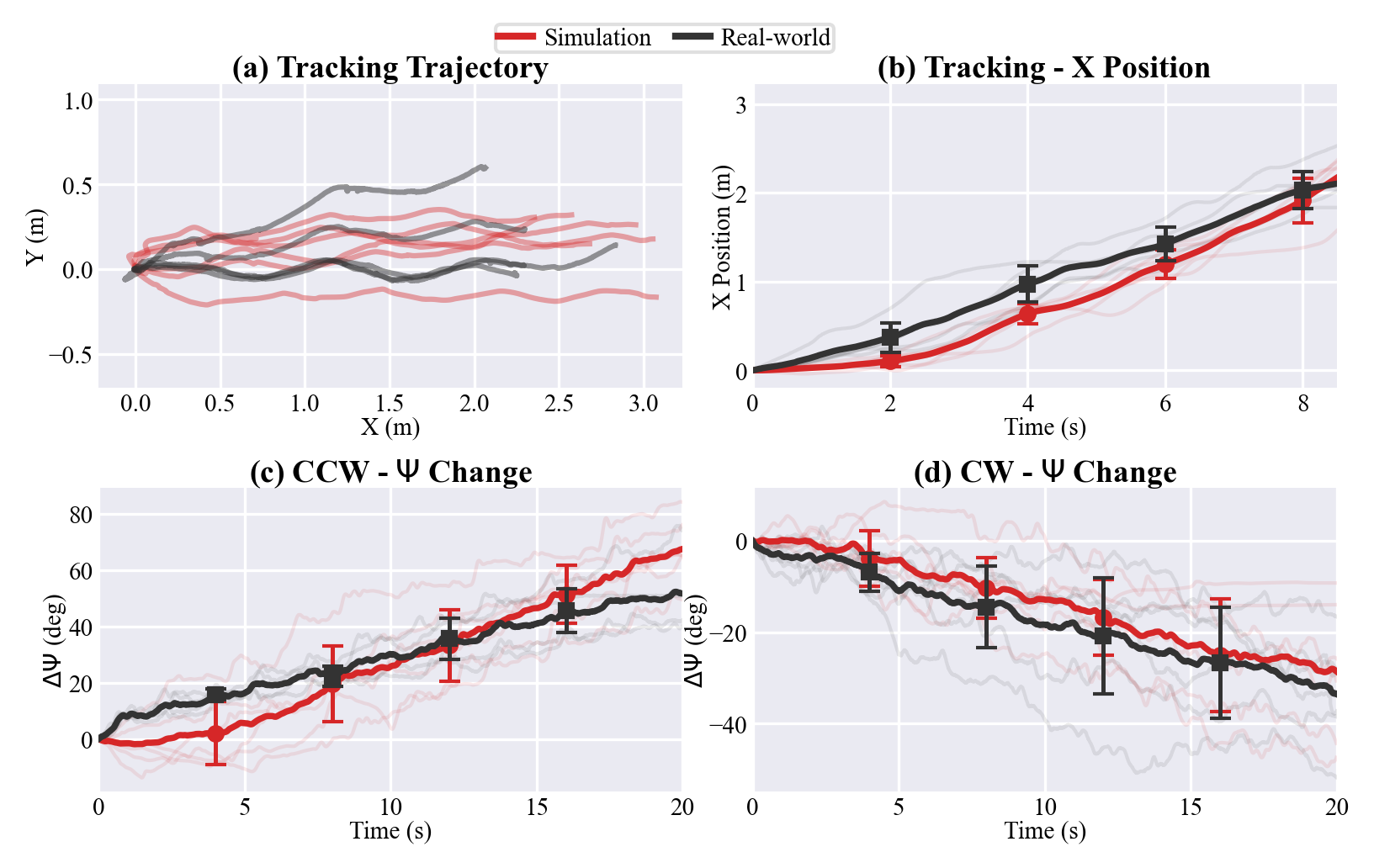}
        \caption{Comparison between simulation and real-world performance. The proposed GNN-SAC demonstrates zero-shot sim-to-real transfer of learned motion primitives, showing comparable behaviors in (a) CoM trajectories, (b) forward tracking, and (c,d) bidirectional turning.}
        \label{fig:sim_real_gap}
    \end{figure}
    
\begin{figure*}[htbp]
        \centering
        \includegraphics[width=\linewidth]{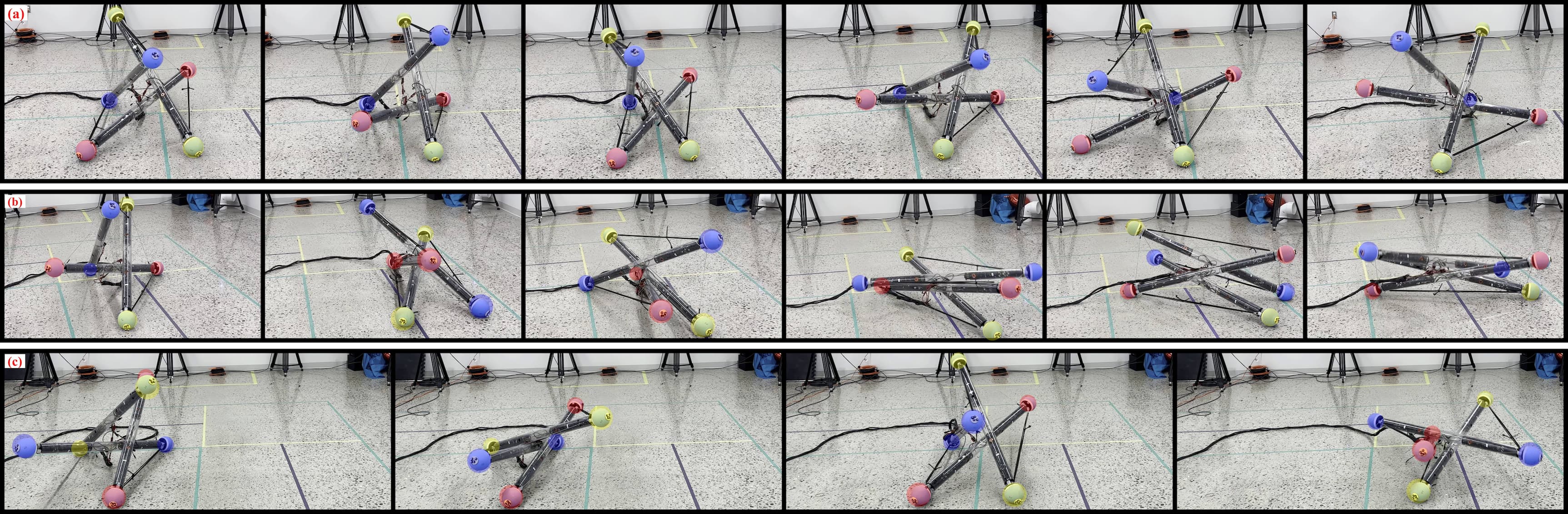}
        \caption{Real-world rollout sequences of learned locomotion primitives. The tensegrity robot executes (a) clockwise turning, (b) counterclockwise turning, and (c) straight-line tracking using zero-shot transferred GNN-SAC policies. The sequences show coordinated rolling and stable motion across all tasks. The colored endcaps indicate node pairs, where yellow, red, and blue correspond to $s_0$-$s_1$, $s_2$-$s_3$, and $s_4$-$s_5$, respectively. See Supplementary Video (0:42-2:13) for the full sequence.}
        \label{fig:real_world}
    \end{figure*}

    
    To enable sim-to-real transfer, the policies trained with G-SAC were deployed on a physical 3-bar tensegrity robot without any additional fine-tuning. All three motion primitives, straight-line tracking, clockwise turning, and counterclockwise turning, were successfully reproduced in real-world trials (Fig.~\ref{fig:real_world}).

    During tracking, the reference point stayed 1~m ahead of the robot; Fig.~\ref{fig:sim_real_gap}(a,b) shows CoM paths and mean forward speed 0.256~m/s before leaving the test area. Counterclockwise turning used a flipping gait (mean $2.76^{\circ}$/s; Fig.~\ref{fig:sim_real_gap}(c)); clockwise turning used coordinated actuation and fore--aft swinging (mean $1.72^{\circ}$/s; Fig.~\ref{fig:sim_real_gap}(d)). Simulation stiffness was $\sim$300~N/m to match the ropes; median sim and real trajectories agree in trend though speeds differ, as primitives were not tuned for velocity tracking.

    The performance gap between simulation and hardware can be attributed to several sources of modeling error. The simulation assumes linear tendon behavior, while the real system exhibits nonlinear stiffness and hysteresis. It also fails to fully capture friction and actuation or sensing delays, causing temporal mismatches.

    The morphology-aware GNN policy helps mitigate these discrepancies by operating on local relational features and promoting distributed coordination through message passing, reducing sensitivity to precise dynamic modeling. This structured inductive bias improves robustness to modeling errors, consistent with the results in Sec.~\ref{sec:robustness_eval}, where stable performance is maintained under variations in stiffness, noise, and terrain.

    Overall, these results confirm that the morphology-aware graph-based policies trained solely in simulation successfully transfer to the physical tensegrity robot in a zero-shot manner, demonstrating stable and coordinated locomotion across all three primitives.

\section{Conclusion}
    In this work, we proposed a morphology-aware graph RL framework for tensegrity robot locomotion by integrating a GNN actor into the Soft Actor-Critic algorithm. By explicitly encoding the robot's structural topology, the GNN-based policy achieved higher sample efficiency, superior final rewards, and stronger robustness than MLP-based baselines. Simulation studies confirmed superior performance on tracking and turning tasks, and robustness under noise, variable stiffness and non-horizontal terrain. Importantly, the learned policies transferred directly to hardware without additional fine-tuning, enabling the 3-bar tensegrity robot to accomplish straight-line tracking and bidirectional turning in real-world experiments. These results demonstrate the effectiveness of morphology-aware policy design for sim-to-real transfer in tensegrity locomotion. In future work, we plan to train a unified policy that can track linear and angular velocities for tensegrity robots.
    
\section*{Acknowledgment}
The author would like to thank Joseph Kennedy for his preliminary work and William Johnson for insightful discussions. The author also gratefully acknowledges Jonathan Mi and Yilin Ma for their hardware support. 





\bibliographystyle{IEEEtran}
\bibliography{ref}

@InProceedings{luo2018tensegrity,
  title={Tensegrity robot locomotion under limited sensory inputs via deep reinforcement learning},
  author={Luo, Jianlan and Edmunds, Riley and Rice, Franklin and Agogino, Alice M},
  booktitle={2018 IEEE International Conference on Robotics and Automation (ICRA)},
  pages={6260--6267},
  year={2018},
  organization={IEEE}
}

@article{tong2024tensegrity,
  title={Tensegrity robot proprioceptive state estimation with geometric constraints},
  author={Tong, Wenzhe and Lin, Tzu-Yuan and Mi, Jonathan and Jiang, Yicheng and Ghaffari, Maani and Huang, Xiaonan},
  journal={IEEE Robotics and Automation Letters},
  volume={10},
  number={4},
  pages={4069--4076},
  year={2025},
  publisher={IEEE},
  URL={https://ieeexplore.ieee.org/document/10910166}
}

@article{mi2024design,
  title={Design of a variable stiffness quasi-direct drive cable-actuated tensegrity robot},
  author={Mi, Jonathan and Tong, Wenzhe and Ma, Yilin and Huang, Xiaonan},
  journal={IEEE Robotics and Automation Letters},
  year={2025},
  publisher={IEEE},
  URL={https://ieeexplore.ieee.org/document/11072300},
  doi={10.1109/LRA.2025.3586519}
}

@book{skelton2009tensegrity,
  title={Tensegrity systems},
  author={Skelton, Robert E and De Oliveira, Mauricio C},
  volume={1},
  year={2009},
  publisher={Springer},
  URL={https://link.springer.com/book/10.1007/978-0-387-74242-7}
}

@article{paul2006design,
  author={Paul, C. and Valero-Cuevas, F.J. and Lipson, H.},
  journal={IEEE Transactions on Robotics}, 
  title={Design and control of tensegrity robots for locomotion}, 
  year={2006},
  volume={22},
  number={5},
  pages={944-957},
  doi={10.1109/TRO.2006.878980}}

@InProceedings{paul2005gait,
  author={Paul, C. and Roberts, J.W. and Lipson, H. and Valero Cuevas, F.J.},
  booktitle={2005 International Conference on Advanced Robotics (ICAR)}, 
  title={Gait production in a tensegrity based robot}, 
  year={2005},
  volume={},
  number={},
  pages={216-222},
  doi={10.1109/ICAR.2005.1507415}}

@InProceedings{kim2015robust,
  author={Kim, Kyunam and Agogino, Adrian K. and Toghyan, Aliakbar and Moon, Deaho and Taneja, Laqshya and Agogino, Alice M.},
  booktitle={2015 IEEE/RSJ International Conference on Intelligent Robots and Systems (IROS)}, 
  title={Robust learning of tensegrity robot control for locomotion through form-finding}, 
  year={2015},
  volume={},
  number={},
  pages={5824-5831},
  doi={10.1109/IROS.2015.7354204}}

@article{kim2020rolling,
author = {Kim, Kyunam and Agogino, Adrian K. and Agogino, Alice M.},
title = {Rolling Locomotion of Cable-Driven Soft Spherical Tensegrity Robots},
journal = {Soft Robotics},
volume = {7},
number = {3},
pages = {346-361},
year = {2020},
doi = {10.1089/soro.2019.0056},
note ={PMID: 32031916},
URL = {https://doi.org/10.1089/soro.2019.0056},
eprint = {https://doi.org/10.1089/soro.2019.0056}
}

@InProceedings{guo2021full,
  author={Guo, Yaqi and Peng, Haijun},
  booktitle={2021 5th International Conference on Robotics and Automation Sciences (ICRAS)}, 
  title={Full-Actuation Rolling Locomotion with Tensegrity Robot via Deep Reinforcement Learning}, 
  year={2021},
  volume={},
  number={},
  pages={51-55},
  doi={10.1109/ICRAS52289.2021.9476651}}

@InProceedings{wang2023real,
  author={Wang, Kun and Johnson, William R. and Lu, Shiyang and Huang, Xiaonan and Booth, Joran and Kramer-Bottiglio, Rebecca and Aanjaneya, Mridul and Bekris, Kostas},
  booktitle={2023 IEEE/RSJ International Conference on Intelligent Robots and Systems (IROS)}, 
  title={Real2Sim2Real Transfer for Control of Cable-Driven Robots Via a Differentiable Physics Engine}, 
  year={2023},
  volume={},
  number={},
  pages={2534-2541},
  doi={10.1109/IROS55552.2023.10341811}}

@InProceedings{cera2018multi,
  author={Cera, Brian and Agogino, Alice M.},
  booktitle={2018 IEEE/RSJ International Conference on Intelligent Robots and Systems (IROS)}, 
  title={Multi-Cable Rolling Locomotion with Spherical Tensegrities Using Model Predictive Control and Deep Learning}, 
  year={2018},
  volume={},
  number={},
  pages={1-9},
  doi={10.1109/IROS.2018.8594401}}

@InProceedings{
    wang2018nervenet,
    title={NerveNet: Learning Structured Policy with Graph Neural Networks},
    author={Tingwu Wang and Renjie Liao and Jimmy Ba and Sanja Fidler},
    booktitle={International Conference on Learning Representations (ICLR)},
    year={2018},
    url={https://openreview.net/forum?id=S1sqHMZCb},
}

@misc{
    haarnoja2018soft,
    title={Soft Actor-Critic: Off-Policy Maximum Entropy Deep Reinforcement Learning with a Stochastic Actor},
    author={Tuomas Haarnoja and Aurick Zhou and Pieter Abbeel and Sergey Levine},
    year={2018},
    url={https://openreview.net/forum?id=HJjvxl-Cb},
}

@misc{
    schulman2017proximal,
    title={Proximal Policy Optimization Algorithms}, 
    author={John Schulman and Filip Wolski and Prafulla Dhariwal and Alec Radford and Oleg Klimov},
    year={2017},
    eprint={1707.06347},
    archivePrefix={arXiv},
    primaryClass={cs.LG},
    url={https://arxiv.org/abs/1707.06347}, 
}

@InProceedings{zhang2017deep,
  author={Zhang, Marvin and Geng, Xinyang and Bruce, Jonathan and Caluwaerts, Ken and Vespignani, Massimo and SunSpiral, Vytas and Abbeel, Pieter and Levine, Sergey},
  booktitle={2017 IEEE International Conference on Robotics and Automation (ICRA)}, 
  title={Deep reinforcement learning for tensegrity robot locomotion}, 
  year={2017},
  volume={},
  number={},
  pages={634-641},
  doi={10.1109/ICRA.2017.7989079}}

@misc{surovik2018adaptive,
      title={Adaptive Tensegrity Locomotion on Rough Terrain via Reinforcement Learning}, 
      author={David Surovik and Kun Wang and Kostas E. Bekris},
      year={2018},
      eprint={1809.10710},
      archivePrefix={arXiv},
      primaryClass={cs.RO},
      url={https://arxiv.org/abs/1809.10710}, 
}

@article{surovik2021adaptive,
    author = {David Surovik and Kun Wang and Massimo Vespignani and Jonathan Bruce and Kostas E Bekris},
    title ={Adaptive tensegrity locomotion: Controlling a compliant icosahedron with symmetry-reduced reinforcement learning},
    journal = {The International Journal of Robotics Research},
    volume = {40},
    number = {1},
    pages = {375-396},
    year = {2021},
    doi = {10.1177/0278364919859443},
    URL = {https://doi.org/10.1177/0278364919859443},
    eprint = {https://doi.org/10.1177/0278364919859443},
}

@article{chen2024learning,
  title={Learning Differentiable Tensegrity Dynamics using Graph Neural Networks},
  author={Chen, Nelson and Wang, Kun and Johnson III, William R and Kramer-Bottiglio, Rebecca and Bekris, Kostas and Aanjaneya, Mridul},
  journal={arXiv preprint arXiv:2410.12216},
  year={2024}
}

@InProceedings{fujimoto2018addressing,
  title={Addressing function approximation error in actor-critic methods},
  author={Fujimoto, Scott and Hoof, Herke and Meger, David},
  booktitle={International Conference on Machine Learning},
  pages={1587--1596},
  year={2018},
  organization={PMLR}
}

@InProceedings{xiong2023universal,
  title={Universal morphology control via contextual modulation},
  author={Xiong, Zheng and Beck, Jacob and Whiteson, Shimon},
  booktitle={International Conference on Machine Learning},
  pages={38286--38300},
  year={2023},
  organization={PMLR}
}

@InProceedings{wang2021sim2sim,
  title={Sim2sim evaluation of a novel data-efficient differentiable physics engine for tensegrity robots},
  author={Wang, Kun and Aanjaneya, Mridul and Bekris, Kostas},
  booktitle={2021 IEEE/RSJ International Conference on Intelligent Robots and Systems (IROS)},
  pages={1694--1701},
  year={2021},
  organization={IEEE}
}

@article{shah2022tensegrity,
  title={Tensegrity robotics},
  author={Shah, Dylan S and Booth, Joran W and Baines, Robert L and Wang, Kun and Vespignani, Massimo and Bekris, Kostas and Kramer-Bottiglio, Rebecca},
  journal={Soft Robotics},
  volume={9},
  number={4},
  pages={639--656},
  year={2022},
  publisher={Mary Ann Liebert, Inc., publishers 140 Huguenot Street, 3rd Floor New~…},
  URL={https://www.liebertpub.com/doi/epub/10.1089/soro.2020.0170}
}

@InProceedings{Tobin2017Domain,
  author={Tobin, Josh and Fong, Rachel and Ray, Alex and Schneider, Jonas and Zaremba, Wojciech and Abbeel, Pieter},
  booktitle={2017 IEEE/RSJ International Conference on Intelligent Robots and Systems (IROS)}, 
  title={Domain randomization for transferring deep neural networks from simulation to the real world}, 
  year={2017},
  volume={},
  number={},
  pages={23-30},
  doi={10.1109/IROS.2017.8202133}}

@InProceedings{Gandhi2017Learning,
  author={Gandhi, Dhiraj and Pinto, Lerrel and Gupta, Abhinav},
  booktitle={2017 IEEE/RSJ International Conference on Intelligent Robots and Systems (IROS)}, 
  title={Learning to fly by crashing}, 
  year={2017},
  volume={},
  number={},
  pages={3948-3955},
  doi={10.1109/IROS.2017.8206247}}

@article{Hwangbo2019Learning,
   title={Learning agile and dynamic motor skills for legged robots},
   volume={4},
   ISSN={2470-9476},
   url={http://dx.doi.org/10.1126/scirobotics.aau5872},
   DOI={10.1126/scirobotics.aau5872},
   number={26},
   journal={Science Robotics},
   publisher={American Association for the Advancement of Science (AAAS)},
   author={Hwangbo, Jemin and Lee, Joonho and Dosovitskiy, Alexey and Bellicoso, Dario and Tsounis, Vassilios and Koltun, Vladlen and Hutter, Marco},
   year={2019},
   month=jan }

@article{andrychowicz2020learning,
  title={Learning dexterous in-hand manipulation},
  author={OpenAI: Andrychowicz, Marcin and Baker, Bowen and Chociej, Maciek and Jozefowicz, Rafal and McGrew, Bob and Pachocki, Jakub and Petron, Arthur and Plappert, Matthias and Powell, Glenn and Ray, Alex and others},
  journal={The International Journal of Robotics Research},
  volume={39},
  number={1},
  pages={3--20},
  year={2020},
  publisher={SAGE Publications Sage UK: London, England}
}

@article{Mnih2015Human,
  author    = {Volodymyr Mnih and Koray Kavukcuoglu and David Silver and Andrei A. Rusu and Joel Veness and Marc G. Bellemare and Alex Graves and Martin Riedmiller and Andreas K. Fidjeland and Georg Ostrovski and Stig Petersen and Charles Beattie and Amir Sadik and Ioannis Antonoglou and Helen King and Dharshan Kumaran and Daan Wierstra and Shane Legg and Demis Hassabis},
  title     = {Human-level control through deep reinforcement learning},
  journal   = {Nature},
  year      = {2015},
  volume    = {518},
  number    = {7540},
  pages     = {529--533},
  doi       = {10.1038/nature14236},
  url       = {https://doi.org/10.1038/nature14236},
  issn      = {1476-4687}
}

@article{lillicrap2015continuous,
  title={Continuous control with deep reinforcement learning},
  author={Lillicrap, Timothy P and Hunt, Jonathan J and Pritzel, Alexander and Heess, Nicolas and Erez, Tom and Tassa, Yuval and Silver, David and Wierstra, Daan},
  journal={arXiv preprint arXiv:1509.02971},
  year={2015}
}

@article{whitman2023learningmodularrobotcontrol,
  author={Whitman, Julian and Travers, Matthew and Choset, Howie},
  journal={IEEE Transactions on Robotics}, 
  title={Learning Modular Robot Control Policies}, 
  year={2023},
  volume={39},
  number={5},
  pages={4095-4113},
  doi={10.1109/TRO.2023.3284362}
  }

@misc{suzuki2026embeddingmorphologytransformerscrossrobot,
      title={Embedding Morphology into Transformers for Cross-Robot Policy Learning}, 
      author={Kei Suzuki and Jing Liu and Ye Wang and Chiori Hori and Matthew Brand and Diego Romeres and Toshiaki Koike-Akino},
      year={2026},
      eprint={2603.00182},
      archivePrefix={arXiv},
      primaryClass={cs.RO},
      url={https://arxiv.org/abs/2603.00182}, 
}

@InProceedings{
    hong2022structureaware,
    title={Structure-Aware Transformer Policy for Inhomogeneous Multi-Task Reinforcement Learning},
    author={Sunghoon Hong and Deunsol Yoon and Kee-Eung Kim},
    booktitle={International Conference on Learning Representations (ICLR)},
    year={2022},
    url={https://openreview.net/forum?id=fy_XRVHqly}
}

\end{document}